%% file: main.tex
\documentclass[letterpaper]{article} 
\usepackage[preprint]{aaai2027}
\usepackage[hyphens]{url} 
\usepackage{graphicx} 
\usepackage{natbib} 
\usepackage{caption} 
\usepackage{booktabs}
\usepackage{colortbl}
\usepackage{multirow}
\usepackage{amsmath}
\usepackage{amssymb}
\usepackage{pifont}
\newcommand{\inlineitem}[1]{\mbox{\raisebox{-0.05ex}{\scalebox{1.08}{\ding{\numexpr171+#1\relax}}}}\nobreakspace}
\newcommand{\takeaway}[1]{\par\smallskip\noindent\begingroup\setlength{\fboxsep}{3pt}\setlength{\fboxrule}{0.35pt}\fcolorbox{gray!35}{gray!8}{\parbox{\dimexpr\linewidth-2\fboxsep-2\fboxrule\relax}{\small\textbf{Takeaway.} #1}}\endgroup\par\smallskip}
\title{SemNav: Semantic Navigation for Repository-Level Issue Localization}
\author{
    Yunxiang Wei\textsuperscript{\rm 1},
    Zhenyu Lei\textsuperscript{\rm 2},
    Jundong Li\textsuperscript{\rm 2}
}
\affiliations{
    \textsuperscript{\rm 1}Independent Researcher\\
    \textsuperscript{\rm 2}University of Virginia\\
    yunxiangwei7@gmail.com\\
    vjd5zr@virginia.edu, jundong@virginia.edu
}

\begin{document}

\maketitle

\begin{abstract}
Repository-level issue localization aims to identify and rank the files and functions relevant to resolving a reported issue. LLM agents approach this task iteratively: they identify a set of potentially relevant locations, inspect the corresponding code, and revise their judgments about these candidates as new evidence is acquired. Existing environments, however, provide limited support for this loop: agents must search for unresolved relation targets, reconstruct entity semantics from raw source code, and revise candidates without evidential basis. To address these limitations, we present SemNav, a framework that leverages deterministic retrieval to seed a broad candidate set and an LLM agent to continually refine that set, thereby combining initial coverage with evidence-guided revision. SemNav supports this process through three key components. A Semantic Navigation Graph resolves program relations on demand through a language server, enabling direct navigation to related entities across files. Issue-conditioned Semantic Cards provide compact, source-grounded interpretations of each entity's role and relevance to the issue. A persistent Candidate Workspace records each candidate together with its evidential basis, enabling grounded verification, revision, and ranking. Across SWE-bench Lite and PLocBench, SemNav outperforms existing baselines, improving File Hit@10 from 68.33\% to 82.67\% with Gemma 4B. Component ablations and trajectory analysis support the complementary roles of all three components, while Semantic Cards reduce working-context load by 48.2\% relative to full-source reading. SemNav further ranks first on all seven evidence-quality metrics on SWE-Explore and improves downstream issue resolution from 44.00\% to 52.33\%.

\end{abstract}

\input{sections/introduction}

\begin{figure*}[t]
    \centering
    \includegraphics[width=\textwidth]{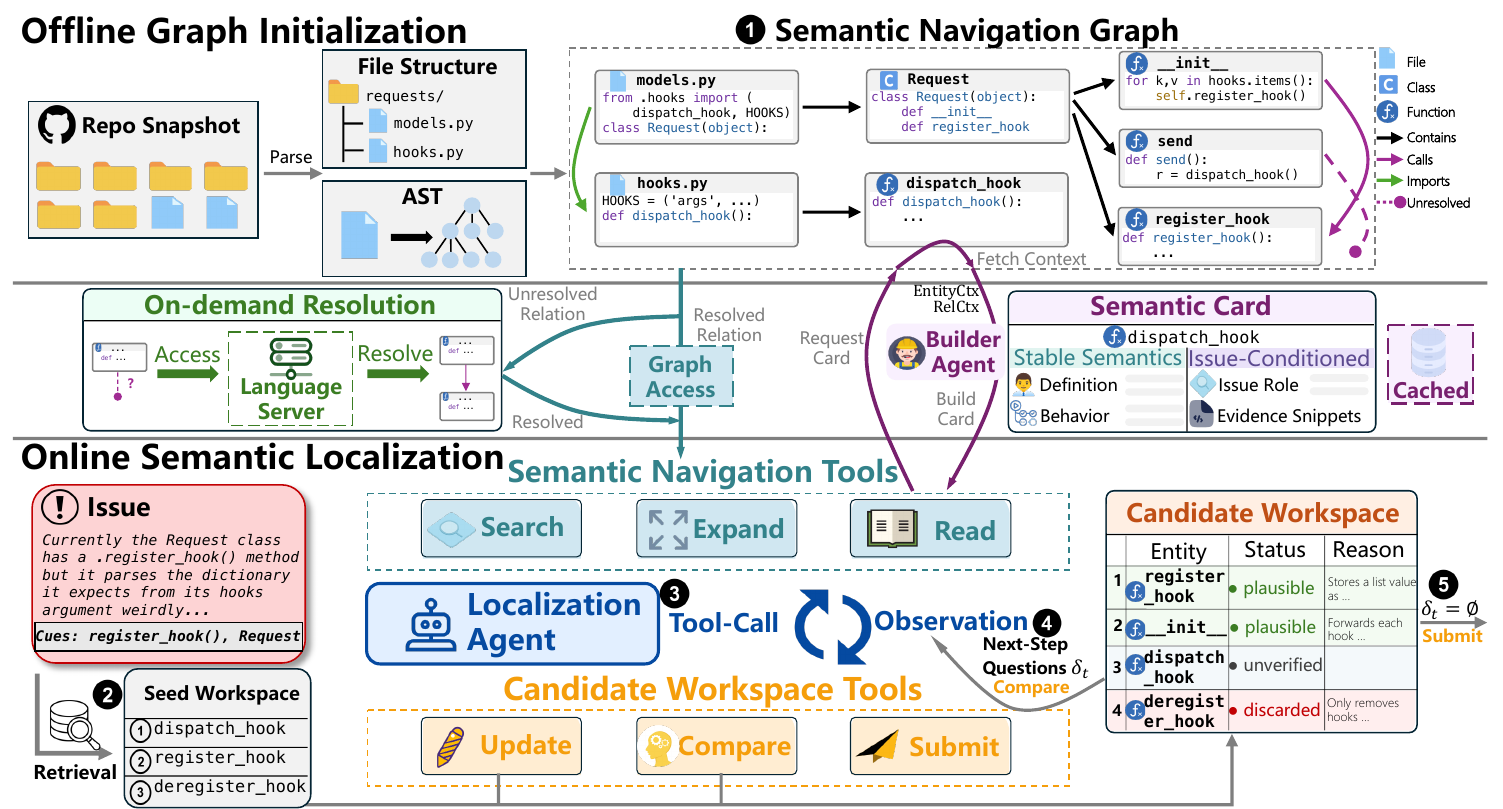}
    \caption{Overview of SemNav.}
    \label{fig:semnav-overview}
\end{figure*}

\input{sections/methodology}
\input{sections/experiments}
\input{sections/related_works}
\section{Conclusion}
We propose \textbf{SemNav}, a framework for repository-level issue localization that combines broad deterministic retrieval with agent-driven investigation. Through a Semantic Navigation Graph, Semantic Cards, and a Candidate Workspace, SemNav makes repository relations, entity interpretations, and evidence-grounded candidate judgments explicit and revisable throughout localization. Experiments across benchmarks and models demonstrate improved localization accuracy and evidence quality, reduced context requirements for the Localization Agent, and better downstream issue resolution.

\bibliography{references}

\end{document}

%% file: sections/introduction.tex
\section{Introduction}

\begin{figure}[t]
    \centering
    \includegraphics[width=0.95\columnwidth]{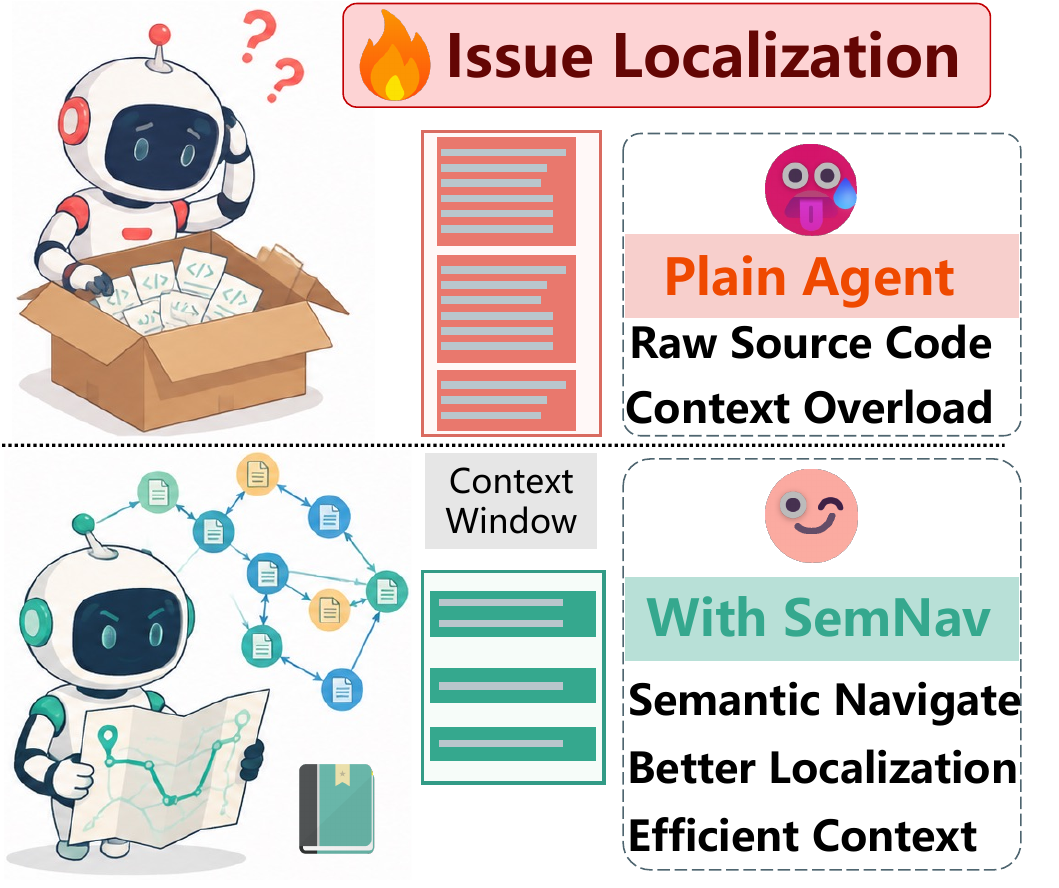}
    \caption{Issue localization with raw-source browsing versus SemNav's semantic navigation and focused repository evidence.}
    \label{fig:semnav-teaser}
\end{figure}

Coding agents are increasingly deployed on real-world software repositories: given a natural-language issue report and a codebase, an agent explores the repository and returns a patch intended to resolve the issue. These repositories often contain far more code than an agent can inspect, while only a small portion is relevant to any single issue, so the agent must first narrow its exploration to a set of potentially relevant locations. \emph{Issue Localization} formalizes this step as the task of identifying and ranking the candidate files and functions that may be involved in resolving the issue. These locations determine which parts of the codebase are available for the agent's subsequent reasoning and editing, and strongly influence downstream issue resolution~\cite{zhang2026sweexplore,sutawika2026codescout}. Effective localization therefore requires high recall over the locations needed for the fix, while keeping the amount of irrelevant code the agent must inspect small. Figure~\ref{fig:semnav-teaser} previews the contrast between raw-source browsing and SemNav's focused repository investigation.

Existing localization methods differ mainly in how they organize this exploration. Workflow-driven methods apply predefined retrieval, filtering, and reranking stages, typically narrowing from repositories to files and then to functions~\cite{xia2025demystifying,xie2025swefixer}. Such pipelines offer broad coverage and a predictable procedure, but pruning is one-way: a location discarded at an early stage cannot be recovered later. Agent-driven methods instead let an LLM decide what to search, read, or navigate next, through general-purpose commands or specialized repository tools~\cite{zhang2024autocoderover,chen2025locagent,zhang2026onetool}. They can acquire repository evidence adaptively, but the locations under consideration and their supporting rationales remain tied to a single path-dependent reasoning trajectory. To retain information beyond the trajectory, recent agent-based methods maintain explicit external state, most commonly a running set of candidate locations~\cite{jiang2025issuelocalization,yu2025orcaloca,ma2025alibaba}. This makes the two designs complementary: a workflow can supply broad initial coverage, while an agent revises that set through continued repository interaction. Localization then becomes a loop in which the current candidates guide evidence acquisition, and newly acquired evidence guides candidate refinement.


We observe that this localization process involves three core actions: acquiring repository evidence, interpreting its relevance to the issue, and revising localization candidates as evidence accumulates. Existing localization environments face unique challenges in supporting these actions. \emph{\textbf{(i) Ambiguous relation targets.}} Syntax-based graphs can identify program relations but cannot always resolve the repository entities they reach; \texttt{obj.run()}, for example, indicates a call without determining which implementation of \texttt{run} is invoked. The agent must therefore search among possible destinations instead of directly navigating connected evidence across files, making evidence acquisition costly. \emph{\textbf{(ii) Implicit entity semantics.}} Although source code contains the information needed to judge relevance, an entity's meaning is often distributed across implementation details rather than stated in compact semantic representation. An agent that consumes raw code must spend its limited context and reasoning capacity reconstructing that purpose before it can interpret the entity against the issue. \emph{\textbf{(iii) Underspecified candidate state.}} A location-only set records which entities are under consideration but not the semantic and evidential basis of its current localization judgment, making successive updates difficult to ground and compare. Together, these limitations place the burden of turning repository observations into evidence-grounded localization decisions on the agent's limited reasoning context.


Software-engineering practice suggests how this burden can be shifted from the reasoner to its environment. When developers encounter unfamiliar code, IDE operations such as go-to-definition take them directly to related implementations, while declarations and documentation help them understand an entity before inspecting its details. Applied to iterative localization, this practice motivates an environment that makes relations directly navigable and entity meaning readily accessible, while preserving candidate judgments as evidence accumulates.


Motivated by this insight, we propose \textbf{SemNav}, an agent-based framework for candidate-centered issue localization. Given an issue and a repository, SemNav leverages deterministic retrieval to obtain a broad set of candidates and then employs an LLM-based Localization Agent to investigate the repository and continually revise them. As illustrated in Figure~\ref{fig:semnav-overview}, SemNav equips the Agent with six tools for navigating repository evidence and revising candidates, supported by three key components. 
\emph{\textbf{(i) Semantic Navigation Graph (SNG).}} The SNG organizes repository entities and resolves requested relation targets online through a language server, allowing the Agent to follow key cross-file relations rather than search among ambiguous destinations. \emph{\textbf{(ii) Semantic Cards.}} On-demand, source-grounded Semantic Cards provide the Agent with a compact entity-level interpretation, reducing the context and reasoning required to recover meaning from implementation details alone. \emph{\textbf{(iii) Candidate Workspace.}} The Candidate Workspace maintains the Agent's current assessment of each candidate together with its evidential basis, allowing candidate revision to remain explicit and grounded as new evidence arrives. Together, these components enable semantic navigation by integrating repository semantics into both the Agent's exploration and its continual revision of candidate judgments.

Experiments on SWE-bench Lite~\cite{jimenez2024swebench} and PLocBench~\cite{zhang2025benchmark} show that SemNav outperforms existing methods across benchmarks and models. With Gemma4 4B, SemNav raises File Hit@10 on SWE-bench Lite from 68.33\% to 82.67\%. Component ablations and trajectory analysis further verify the complementary roles of our designed three components: the online relation resolution and the Semantic Cards improve localization accuracy, the Candidate Workspace improves both top-ranked accuracy and candidate coverage during iterative revision, and the Semantic Cards reduce the context required to interpret repository entities. On SWE-Explore~\cite{zhang2026sweexplore}, SemNav ranks first on all seven evidence-quality metrics, raising line-level precision from 53.35\% to 79.99\%. Finally, when the localized evidence is used as the repository context for program repair, SemNav achieves the highest patch application and issue-resolution rates, increasing the resolution rate from 44.00\% to 52.33\% with the Qwen3.5 35B repair backend.

%% file: sections/methodology.tex
\section{Methodology}

\subsection{Background}
\label{sec:method-background}

An abstract syntax tree (AST) represents source code through grammatical constructs such as declarations, expressions, and statements. Its hierarchy records how these constructs are nested, while source spans map them back to their locations in a file. AST traversal can therefore identify repository entities and locate syntactic occurrences of calls, imports, references, and inheritance. However, syntax records the form and location of an occurrence rather than necessarily determining the project-wide binding of the symbol it contains.

A language server performs project-aware static analysis and exposes position-indexed code-intelligence operations. Go-to-definition maps a symbol occurrence to its resolved definition when available, whereas find-references returns source positions associated with a given definition. Our Python implementation uses Pyright\footnote{\url{https://microsoft.github.io/pyright/}} for these operations. Because their results depend on statically available project information, language-server queries may remain unresolved for dynamic behavior or insufficiently constrained symbols.

\subsection{Overview}
\label{sec:method-overview}

Given a repository snapshot $\mathcal R$, a natural-language issue $q$, and a target granularity $\tau\in\{\mathrm{file},\mathrm{func}\}$, repository-level issue localization aims to return an ordered list of $K$ code entities implicated in resolving the issue:
\begin{equation}
    \widehat{\mathbf L}^{\tau}_q
    =\left\langle v_1,\ldots,v_K\right\rangle,
    \qquad v_i\in\mathcal V_{\tau}.
    \label{eq:localization-task}
\end{equation}
Here, $\mathcal V_{\tau}$ is the set of file or function targets; class entities are retained as intermediate navigation structures. For each issue, a lightweight rule-based parser identifies explicit code mentions in $q$ and normalizes them into a set of source-facing cues, denoted by $\phi(q)$, such as file paths, identifiers, and API calls.

The SemNav pipeline consists of two stages. In \emph{offline graph initialization}, SemNav uses AST analysis to initialize the Semantic Navigation Graph (SNG) and builds an entity-level retrieval index, retaining relations whose targets remain unresolved. In \emph{online semantic localization}, a deterministic retrieval pass uses $q$ and $\phi(q)$ to initialize the Candidate Workspace $W_0$. An LLM-based \emph{Localization Agent} then investigates the repository through the SNG while maintaining and revising localization hypotheses in the Workspace. During this investigation, SemNav queries the language server to resolve relation targets on demand and updates the SNG with the successfully resolved relations.

We equip the Localization Agent with six tools organized into two groups. The first group comprises the \emph{semantic navigation tools} \textsc{Search}, \textsc{Expand}, and \textsc{Read}, which allow the Agent to locate entities, access program relations through the SNG, and inspect source code or on-demand Semantic Cards. The second comprises the \emph{Candidate Workspace tools} \textsc{Update}, \textsc{Compare}, and \textsc{Submit}, which record candidate judgments, compare plausible candidates, and produce the final ranking. The navigation tools return observations without modifying the Workspace; \textsc{Update} changes candidate records, while \textsc{Compare} may reorder them before submission. Starting from $W_0$, the Agent alternates between semantic navigation and Workspace updates. When invoked, \textsc{Compare} either returns a focused verification question $\delta_t$ that guides further investigation or confirms the current Workspace, after which \textsc{Submit} returns the ranked locations in Eq.~\eqref{eq:localization-task}. The following subsections detail the SNG, semantic navigation with Semantic Cards, and the Candidate Workspace, respectively.

\subsection{Semantic Navigation Graph}
\label{sec:graph-construction}

Given a repository $\mathcal R$, SemNav represents its current SNG as
\begin{equation}
    \mathcal G
    =\left(\mathcal V,\mathcal E,\widetilde{\mathcal E}\right),
    \qquad
    \mathcal V
    =\mathcal V_{\mathrm{file}}\cup\mathcal V_{\mathrm{class}}\cup\mathcal V_{\mathrm{func}},
    \label{eq:sng-state}
\end{equation}
where $\mathcal E\subseteq\mathcal V\times\mathcal T\times\mathcal V\times\mathcal L$ contains resolved edges and $\widetilde{\mathcal E}\subseteq\mathcal V\times\mathcal T\times\mathcal L$ contains unresolved edges. Here, $\mathcal L$ is the set of source locations. A resolved edge $e=(u,r,v,\ell)$ represents relation $r$ from entity $u$ to entity $v$ at location $\ell$. An unresolved edge $\widetilde e=(u,r,\ell)$ leaves its target to online resolution. Methods are represented as function entities, and each entity is identified by its qualified name and source span. For example, \texttt{func:pkg/core.py::Parser.parse} identifies the method \texttt{Parser.parse} in \texttt{pkg/core.py}.

The relation set $\mathcal T$ contains \textsc{Contains}, \textsc{Calls}, \textsc{Imports}, \textsc{Inherits}, and \textsc{References}. \textsc{Contains} connects a file or class to a directly declared entity; \textsc{Calls} connects a caller to its callee; \textsc{Imports} connects an importing entity to an imported repository entity; and \textsc{Inherits} connects a subclass to its base class. \textsc{References} covers the remaining symbol uses that are not categorized as calls, imports, or inheritance. All relations are directed from the source entity to the target entity.

Offline graph initialization constructs
\begin{equation}
    \mathcal G_0
    =\left(\mathcal V,\mathcal E_0,\widetilde{\mathcal E}_0\right)
    =\operatorname{ASTInit}(\mathcal R).
    \label{eq:ast-extraction}
\end{equation}
AST analysis extracts repository entities and identifies the source positions of relevant syntactic constructs, which we call \emph{relation sites}. Relations whose targets are syntactically determined form $\mathcal E_0$, while relation sites with unresolved targets are recorded in $\widetilde{\mathcal E}_0$.

During online localization, SemNav resolves the relations requested for an entity through
\begin{equation}
    \mathcal G'
    =\operatorname{Access}\!\left(\mathcal G;v,r,d\right),
    \qquad d\in\{+,-\}.
    \label{eq:relation-access}
\end{equation}
\textsc{Access} retrieves resolved relations from $\mathcal E$ and queries the language server to resolve the relevant relation sites recorded in $\widetilde{\mathcal E}$. For forward access ($d=+$), it applies go-to-definition at each outgoing unresolved edge $\widetilde e=(v,r,\ell)$; a successfully resolved target $v'$ produces $e=(v,r,v',\ell)$. For reverse access ($d=-$), it applies find-references to $v$ and matches the returned relation sites to unresolved edges $\widetilde e=(u,r,\ell)$, producing $e=(u,r,v,\ell)$. Successfully resolved edges are added to $\mathcal E$, while sites without a resolved repository target remain in $\widetilde{\mathcal E}$, yielding the updated SNG $\mathcal G'$.

\subsection{Semantic Navigation and Semantic Cards}
\label{sec:semantic-navigation}

SemNav exposes the SNG through three evidence-facing tools. \textsc{Search} establishes issue-relevant entry points, \textsc{Expand} follows typed relations between entities, and \textsc{Read} presents selected entities at structural, source, and semantic levels. Whenever a tool requires program relations, SemNav applies \textsc{Access} to resolve the relevant edges and operates on the updated SNG $\mathcal G'$.

\paragraph{Search.}
For each entity $v$, SemNav indexes a source-grounded textual representation $D(v)$ formed from its qualified name, declaration, documentation, and implementation text. Given an Agent-generated query $x$, issue cues $\phi(q)$, and an optional entity or path scope $\sigma$, \textsc{Search} first computes BM25 lexical relevance~\cite{robertson2009probabilistic} and selects an anchor set:
\begin{equation}
\begin{aligned}
    b(v\mid x,q)
        &=\alpha\operatorname{BM25}\!\left(x,D(v)\right)
          +\beta\sum_{c\in\phi(q)}\mu(c,v),\\
    \mathcal A
        &=\underset{u\in\mathcal V_{\sigma}}{\operatorname{Top}_{L}}\,
          b(u\mid x,q).
    \label{eq:search-anchors}
\end{aligned}
\end{equation}
Here, the matcher $\mu(c,v)\in\{0,1\}$ indicates whether a normalized cue $c$ has an exact keyword or source-location match in $v$. Search then reranks entities using the resolved neighborhoods of the anchors in the updated SNG $\mathcal G'$:
\begin{equation}
    \operatorname{Search}(x,\sigma;q)
    =\underset{v\in\mathcal V_{\sigma}}{\operatorname{Top}_{M}}
      \left[b(v\mid x,q)
      +\gamma\sum_{u\in\mathcal A}\mathbf{1}
      \left[u\sim_{\mathcal G'}v\right]\right].
    \label{eq:entity-search}
\end{equation}
We write $u\sim_{\mathcal G'}v$ when a resolved SNG edge connects $u$ and $v$ in either direction. \textsc{Search} returns the top $M$ entities together with the textual, cue, and relation evidence supporting each match.

\paragraph{Expand.}
\textsc{Expand} performs typed one-hop navigation over the SNG. For relation type $r\in\mathcal T$, the outgoing and incoming neighborhoods in a current graph state are
\begin{equation}
\begin{aligned}
    \mathcal N_r^{+}(v;\mathcal G)
        &=\{u\mid\exists\ell:(v,r,u,\ell)\in\mathcal E\},\\
    \mathcal N_r^{-}(v;\mathcal G)
        &=\{u\mid\exists\ell:(u,r,v,\ell)\in\mathcal E\}.
    \label{eq:relation-neighborhoods}
\end{aligned}
\end{equation}
\begin{equation}
    \operatorname{Expand}(v,r,d)
        =\mathcal N_r^d(v;\mathcal G'),
        \qquad d\in\{+,-\}.
    \label{eq:entity-expand}
\end{equation}
\textsc{Expand} returns the resulting neighbors together with their relation types, directions, and source locations.

\paragraph{Read.}
\textsc{Read} presents a selected entity at three evidence levels:
\begin{equation}
    \operatorname{Read}(v,m;q)=
    \begin{cases}
        \operatorname{Summary}(v), & m=\mathrm{summary},\\
        \operatorname{Code}(v), & m=\mathrm{code},\\
        C_q(v), & m=\mathrm{card}.
    \end{cases}
    \label{eq:entity-read}
\end{equation}
The issue $q$ conditions only the card mode. The summary mode is a deterministic projection of the indexed declaration, containment structure, and immediate relation profile of $v$, providing a compact structural orientation. The code mode returns the original source span without accessing additional relations, providing the richest and most authoritative evidence about the entity. While large spans consume substantial context, they may be truncated under the observation budget and require the Localization Agent to reconstruct their semantics. The card mode directly presents an issue-conditioned semantic understanding together with compact source snippets that support it.

\input{tables/main_results}

\paragraph{Semantic Card.}
A Semantic Card is a source-grounded interpretation of one repository entity. It records \inlineitem{1}the entity's definition and responsibility; \inlineitem{2}its operational behavior; \inlineitem{3}its role in the reported issue; and \inlineitem{4}compact source snippets supporting these interpretations. SemNav constructs the Card with a single Builder Agent invocation:
\begin{equation}
\begin{split}
    X_q(v)
        &=\bigl(q,\operatorname{EntityCtx}(v),
           \operatorname{RelCtx}(v;\mathcal G')\bigr),\\
    C_q(v)
        &=\operatorname{Builder}\!\left(X_q(v)\right),
    \label{eq:semantic-card}
\end{split}
\end{equation}
where $\mathcal G'$ denotes the SNG after accessing the relation families used by the Card. The entity context combines the declaration, implementation, and containment structure of $v$. The relational context collects the resulting typed one-hop neighborhoods,
\begin{equation}
    \operatorname{RelCtx}(v;\mathcal G')
    =\{(u,r,\pm)\mid r\in\mathcal T,\ u\in\mathcal N_r^{\pm}(v;\mathcal G')\},
    \label{eq:relation-context}
\end{equation}
where $+$ and $-$ denote outgoing and incoming relations, respectively. Each compact relation record identifies the neighboring entity together with the edge type and direction; the returned relation evidence also retains its supporting source location. The Builder's dedicated context window holds the target source and relational evidence, while the Localization Agent receives only the compact Card and its supporting snippets.

Cards are constructed on demand and cached by repository snapshot, entity, and issue context, allowing repeated reads to reuse the same semantic interpretation. The Localization Agent consumes $C_q(v)$ as a compact entity-level understanding and can request source code when the investigation requires more detailed implementation evidence.

\subsection{Candidate Workspace}
\label{sec:candidate-workspace}

SemNav organizes the localization process around a persistent \emph{Candidate Workspace}, which maintains the candidates discovered during investigation and their current priority. For a requested target granularity $\tau\in\{\mathrm{file},\mathrm{func}\}$, the Workspace at step $t$ is an ordered list of candidate records:
\begin{equation}
    h_t^i=\bigl(v_t^i,z_t^i,\eta_t^i\bigr),
        \quad v_t^i\in\mathcal V_{\tau},
        \qquad
    W_t=\left\langle h_t^1,\ldots,h_t^{n_t}\right\rangle.
    \label{eq:candidate-workspace}
\end{equation}
The list order gives the current candidate ranking. Each record $h_t^i$ contains the entity $v_t^i$, its investigation status $z_t^i$, and an evidence-grounded rationale $\eta_t^i$ that cites the observations supporting its current assessment. An \emph{unverified} entity has been retrieved or discovered and awaits further examination, a \emph{plausible} entity is supported by the acquired evidence, and a \emph{discarded} entity has been ruled out.

SemNav initializes the Workspace before the Agent begins its investigation:
\begin{equation}
    W_0=\operatorname{Seed}_{\tau}(q,\mathcal G).
    \label{eq:candidate-seed}
\end{equation}
The seeding procedure applies the retrieval mechanism of \textsc{Search} using the issue description as a fixed query. It inserts the retrieved entities as unverified records and uses their retrieval ranking as the initial candidate order. The resulting Workspace provides the Agent with a broad starting point and can be extended with candidates discovered during subsequent investigation.

\paragraph{Update.}
To operate on the Candidate Workspace, SemNav provides the Localization Agent with an explicit \textsc{Update} tool:
\begin{equation}
    W_{t+1}=\operatorname{Update}\bigl(W_t;v,z,\eta\bigr).
    \label{eq:candidate-update}
\end{equation}
The Agent uses \textsc{Update} to insert a newly discovered candidate or revise the status and rationale of an existing record. As evidence accumulates, an unverified candidate can be promoted to plausible or marked as discarded. The Agent can also refine a coarse repository location by adding the concrete target entities reached from it. These updates externalize the Agent's evolving entity-level judgments in a persistent and revisable candidate state.

\paragraph{Compare.}
Before submission, SemNav confirms the current Workspace through \textsc{Compare}:
\begin{equation}
    (\bar W_t,\delta_t)=\operatorname{Compare}(q,W_t).
    \label{eq:candidate-compare}
\end{equation}
\textsc{Compare} invokes a dedicated Comparison Agent with a bounded dossier containing the rationale, source code, and Semantic Card of each plausible candidate in $W_t$. The Comparison Agent jointly evaluates these candidates according to their issue relevance and implementation responsibility and writes their resulting order to $\bar W_t$. The reranked plausible candidates lead $\bar W_t$, while the unverified candidates retain their maintained order. It then assesses whether the leading candidates are sufficiently supported and distinguished for submission.

When further confirmation is required, $\delta_t$ specifies a focused verification question. The Localization Agent uses this question to continue semantic navigation and applies further \textsc{Update} operations to $\bar W_t$, producing the next Workspace state $W_{t+1}$ before invoking \textsc{Compare} again.

Notably, the Builder and Comparison Agents serve as auxiliary agents with separate context windows. They support semantic construction and candidate comparison without carrying their full working contexts into the Localization Agent's trajectory.

\paragraph{Submit.}
The localization loop terminates when \textsc{Compare} confirms the current ranking or when the budget is exhausted; we denote the terminal step by $T$. The Agent then submits the latest compared Workspace:
\begin{equation}
    \widehat{\mathbf L}^{\tau}_q=\operatorname{Submit}_{\tau}(\bar W_T).
    \label{eq:candidate-submit}
\end{equation}
\textsc{Submit} first returns plausible candidates in the confirmed order and, when fewer than $K$ are available, fills the remaining positions with the highest-ranked unverified candidates. Discarded candidates are excluded. This process transforms broad deterministic retrieval into an evidence-grounded ranking that remains revisable throughout the investigation.

%% file: tables/main_results.tex
\begin{table*}[!t]
    \caption{Localization results on SWE-bench Lite and PLocBench. Best and second-best results are bold and underlined.}
\label{tab:main_results}
\centering
\newcommand{\mainresultsoffset}{-0.035\textwidth}
\hspace*{\mainresultsoffset}%
\begin{minipage}[t]{0.57\textwidth}
    \vspace{0pt}
    \centering
    \begingroup
    \renewcommand{\arraystretch}{0.90}
    \scriptsize
    \setlength{\tabcolsep}{1.8pt}
    \begin{tabular}{@{}llrrrrrrrrr@{}}
    \toprule
    \multicolumn{11}{@{}l}{\textbf{(a) SWE-bench Lite}} \\
    \midrule
    & & \multicolumn{3}{c}{File Hit} & \multicolumn{3}{c}{Function Hit} & \multicolumn{3}{c}{Function Recall} \\
    \cmidrule(lr){3-5}\cmidrule(lr){6-8}\cmidrule(l){9-11}
    Model & Method & @1 & @5 & @10 & @1 & @5 & @10 & @1 & @5 & @10 \\
    \midrule
    \multirow{6}{*}{\shortstack[l]{Gemma\\4B}} & Agentless       & 36.00 & 45.67 & 45.67 &  4.67 &  5.00 &  5.00 &  2.83 &  3.11 &  3.11 \\
    & CoSIL           & 51.67 & \underline{68.33} & \underline{68.33} & 24.00 & 31.33 & 31.33 & 16.40 & 22.91 & 22.91 \\
    & LocAgent        & \underline{54.33} & 62.33 & 62.33 & 22.67 & \underline{42.33} & \underline{43.00} & 16.41 & \underline{31.92} & \underline{32.92} \\
    & mini-swe-agent  & 49.00 & 49.33 & 49.33 & \underline{25.33} & 28.00 & 28.00 & \underline{22.28} & 25.11 & 25.11 \\
    & RepoSearcher    & 42.67 & 56.33 & 56.33 &  1.67 &  5.67 &  5.67 &  1.23 &  4.51 &  4.51 \\
    \rowcolor{gray!15}[3pt][3pt] & \textbf{SemNav} & \textbf{58.33} & \textbf{78.00} & \textbf{82.67} & \textbf{34.33} & \textbf{47.67} & \textbf{52.00} & \textbf{28.72} & \textbf{35.40} & \textbf{40.72} \\
    \multirow{6}{*}{\shortstack[l]{Qwen3.5\\9B}} & Agentless       & 44.00 & 58.33 & 58.67 & 12.67 & 28.00 & 31.00 & 12.00 & 25.89 & 29.06 \\
    & CoSIL           & \underline{64.00} & \textbf{80.67} & \underline{80.67} & \underline{42.33} & \textbf{58.33} & \underline{58.33} & \underline{31.97} & 43.45 & 45.84 \\
    & LocAgent        & 59.33 & 74.33 & 74.33 & 24.33 & 51.00 & 54.00 & 22.72 & \underline{48.47} & \underline{51.42} \\
    & mini-swe-agent  & 29.33 & 31.67 & 31.67 & 25.33 & 28.33 & 28.33 & 22.72 & 26.89 & 26.89 \\
    & RepoSearcher    &  6.00 &  7.00 &  7.00 &  0.67 &  1.67 &  1.67 &  0.50 &  1.39 &  1.39 \\
    \rowcolor{gray!15}[3pt][3pt] & \textbf{SemNav} & \textbf{66.00} & \underline{79.67} & \textbf{83.33} & \textbf{47.00} & \underline{57.33} & \textbf{59.00} & \textbf{38.58} & \textbf{55.14} & \textbf{55.14}\\
    \multirow{6}{*}{\shortstack[l]{Qwen3.5\\35B-A3B}} & Agentless       & 65.33 & 86.33 & 86.33 & 21.00 & 54.33 & 57.00 & 19.33 & 52.00 & 54.56 \\
    & CoSIL           & 70.67 & \textbf{89.33} & \underline{89.33} & 39.00 & 48.33 & 48.33 & 35.28 & 45.92 & 45.92 \\
    & LocAgent        & \underline{73.67} & 80.00 & 80.00 & 27.67 & \underline{58.00} & \underline{62.00} & 25.44 & \underline{55.67} & \underline{59.50} \\
    & mini-swe-agent  & 68.00 & 69.00 & 69.00 & \underline{53.00} & 56.00 & 56.33 & \underline{47.00} & 52.89 & 53.22 \\
    & RepoSearcher    & 70.67 & 87.33 & 87.33 &  7.67 & 22.00 & 22.00 &  6.94 & 20.06 & 20.06 \\
    \rowcolor{gray!15}[3pt][3pt] & \textbf{SemNav} & \textbf{78.67} & \underline{88.33} & \textbf{90.00} & \textbf{63.67} & \textbf{72.67} & \textbf{74.00} & \textbf{49.19} & \textbf{56.66} & \textbf{60.70} \\
    \bottomrule
    \end{tabular}
    \endgroup
    \end{minipage}
    \hfill
    \begin{minipage}[t]{0.41\textwidth}
    \vspace{0pt}
    \centering
    \begingroup
    \renewcommand{\arraystretch}{0.92}
    \scriptsize
    \setlength{\tabcolsep}{1.9pt}
    \begin{tabular*}{\linewidth}{@{\extracolsep{\fill}}lrrrrr@{}}
    \toprule
    \multicolumn{6}{@{}l}{\textbf{(b) PLocBench}} \\
    \midrule
    Method & Acc@3 & Acc@5 & MRR@5 & Macro R@5 & Micro R@5 \\
    \midrule
    Agentless       & 46.16 & 47.75 & 40.48 & 32.46 & 22.23 \\
    CoSIL           & \underline{57.73} & \underline{60.41} & 50.78 & \underline{42.07} & \underline{28.65} \\
    LocAgent        & 37.39 & 37.64 & 34.73 & 25.47 & 16.90 \\
    mini-swe-agent  & 56.76 & 56.88 & \underline{56.55} & 37.80 & 21.64 \\
    RepoSearcher    & 54.57 & 57.13 & 48.20 & 39.56 & 27.02 \\
    \rowcolor{gray!15}[3pt][3pt]\textbf{SemNav} & \textbf{63.09} & \textbf{67.36} & \textbf{57.50} & \textbf{47.45} & \textbf{31.77} \\
    \bottomrule
    \end{tabular*}
    \endgroup
    \vspace{2.1em}
    \captionof{table}{Evidence acquisition results on SWE-Explore.}
    \label{tab:swe_explore_results}
    \begingroup
    \renewcommand{\arraystretch}{0.92}
    \scriptsize
    \setlength{\tabcolsep}{1.5pt}
    \begin{tabular*}{\linewidth}{@{\extracolsep{\fill}}lrrrrrrr@{}}
    \toprule
    Method & P & F1 & R@500 & nDCG & HitFile & FUH & CE \\
    \midrule
    Agentless       & 43.34 & 33.35 & 21.46 & 54.88 & 26.84 & 58.79 & 51.20 \\
    CoSIL           & 49.33 & \underline{50.53} & \underline{34.39} & \underline{76.28} & \underline{39.50} & \underline{82.57} & 63.50 \\
    LocAgent        & 53.07 &  9.02 &  5.88 & 55.14 & 31.94 & 58.79 & 59.42 \\
    mini-swe-agent  & \underline{53.35} & 13.55 &  9.28 & 75.77 & 33.19 & 77.79 & \underline{63.98} \\
    \rowcolor{gray!15}[3pt][3pt]\textbf{SemNav} & \textbf{79.99} & \textbf{66.94} & \textbf{37.83} & \textbf{86.96} & \textbf{42.42} & \textbf{86.96} & \textbf{88.32} \\
    \bottomrule
    \end{tabular*}
    \endgroup
    \end{minipage}
    \end{table*}

%% file: sections/experiments.tex
\begin{figure*}[t]
    \centering
    \includegraphics[width=\textwidth]{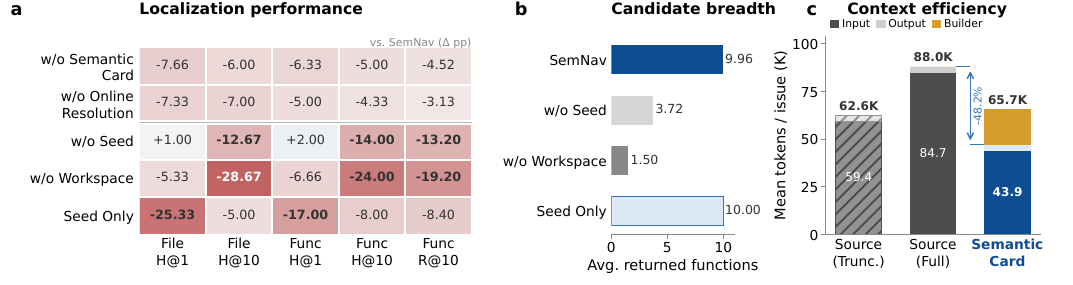}
    \caption{Component analyses on SWE-bench Lite with Gemma 4B.}
    \label{fig:ablation_analysis}
\end{figure*}

\section{Experiments}

\subsection{Experimental Setup}

We evaluate SemNav on three benchmarks. SWE-bench Lite contains 300 issues and evaluates ranked file- and function-level localization~\cite{jimenez2024swebench}. PLocBench contains 821 issues and evaluates ranked file localization~\cite{zhang2025benchmark}. We further use 368 SWE-Explore instances to evaluate the source evidence returned by each method beyond conventional localization targets~\cite{zhang2026sweexplore}.

For the cross-benchmark comparison, all methods use the same \textit{gemma-4-e4b-it} model (Gemma 4B). On SWE-bench Lite, we also evaluate every method with \textit{Qwen3.5 9B} and \textit{Qwen3.5 35B-A3B} to test whether the results hold across model families and scales. We compare SemNav with workflow-driven and agent-driven baselines, including Agentless~\cite{xia2025demystifying}, CoSIL~\cite{jiang2025issuelocalization}, LocAgent~\cite{chen2025locagent}, mini-swe-agent~\cite{sweagentteam2024minisweagent}, and RepoSearcher~\cite{ma2025toolintegrated}.

On SWE-bench Lite, Hit@$K$ measures whether the first $K$ predictions contain a gold entity, while Function Recall@$K$ measures coverage of all gold functions. On PLocBench, Acc@$K$ measures whether the Top-$K$ list contains a gold file, MRR@5 rewards earlier first hits, and Macro/Micro Recall@5 measure gold-file coverage. On SWE-Explore, P and F1 measure line-level precision and F1; Recall@500 and nDCG@500 evaluate evidence coverage and ranking within a 500-line budget; HitFile measures coverage of core files; First Useful Hit (FUH) rewards earlier useful evidence; and Context Efficiency (CE) measures the fraction of predicted lines that belong to core or optional context. All metrics are reported as percentages, and higher values are better.

\subsection{Overall Localization Performance}

SemNav achieves the strongest overall localization performance on both benchmarks. On SWE-bench Lite, it ranks first in 24 of the 27 model--metric settings and second in the remaining three (Table~\ref{tab:main_results}, Panel~(a)). The gains hold with both Gemma and Qwen and at both Qwen model scales. They are most consistent at the function level, where SemNav improves both Function Hit and Function Recall with all three models. For example, with Gemma 4B, SemNav raises Function Recall@10 from 32.92 to 40.72 over the strongest baseline.

The same advantage holds for ranked file localization on PLocBench (Table~\ref{tab:main_results}, Panel~(b)). SemNav ranks first on all five metrics, improving Acc@5 from 60.41 to 67.36 and Macro Recall@5 from 42.07 to 47.45. The joint gains in accuracy, MRR, and recall show that SemNav improves both the ranking and coverage of gold files.

\takeaway{SemNav improves localization across benchmarks, model families, and model scales.}

\subsection{Component Contributions}

We ablate the three main components of SemNav on SWE-bench Lite with Gemma 4B (Figure~\ref{fig:ablation_analysis}, Panels~(a)--(b)). For Semantic Navigation, \emph{w/o Semantic Card} replaces issue-conditioned Cards with the remaining source views, while \emph{w/o Online Resolution} retains only the offline AST-initialized SNG. For the Candidate Workspace, \emph{w/o Seed} starts from an empty Workspace, \emph{w/o Workspace} directly returns the Localization Agent's predictions, and \emph{Seed Only} returns the deterministic initialization without agent investigation.

Both Semantic Cards and online resolution contribute to localization performance. Removing either component lowers all five metrics while leaving the average output size unchanged. File Hit@10 drops by 6.00 points without Semantic Cards and by 7.00 points without online resolution. These drops come from the information available to the agent rather than from returning fewer candidates.

The Seed provides coverage, while agent investigation improves ranking. Removing the Seed slightly improves Hit@1 but lowers Function Recall@10 by 13.20 points. Seed Only preserves a broad candidate set but ranks its head poorly; the full method improves File Hit@1 by 25.33 points through agent investigation. The Workspace keeps candidates available during this process. Without it, the average output drops from 9.96 to 1.50 functions and Function Recall@10 falls by 19.20 points.

We next measure how Semantic Cards change the context processed by the Localization Agent (Figure~\ref{fig:ablation_analysis}, Panel~(c)). We replay the same SemNav trajectories with the same tool actions and observation order, varying only the observations returned by the \textsc{Read} tool. For each \textsc{Read} action, the three variants return truncated source code, full source code, or the Semantic Card of the entity, respectively. By keeping the trajectories fixed, this setup isolates the effect of context representation from the Agent's exploration policy.

Semantic Cards reduce the Localization Agent's working context by 48.2\% relative to full-source reading and also reduce its peak context. Building the Cards requires additional tokens, but the total cost remains 25.4\% below full-source reading and only 4.8\% above source truncation. Semantic Cards therefore improve localization while giving the Localization Agent a more compact working context.

\takeaway{Broad seeding alone is not enough: SemNav uses the Workspace to retain and revise candidates, while Semantic Cards reduce the context needed for these decisions.}

\subsection{Evidence Acquisition Quality}

SemNav returns cleaner localization context on SWE-Explore, ranking first on all seven metrics (Table~\ref{tab:swe_explore_results}). Its largest gains are in precision, F1, and context efficiency: line-level precision rises from 53.35\% to 79.99\%, while Recall@500 also improves. The higher precision therefore does not come from narrowing evidence coverage. Instead, SemNav removes more irrelevant context while preserving the regions needed for localization. This pattern is consistent with its use of resolved relations and issue-conditioned Semantic Cards to guide localization toward relevant code.

\takeaway{SemNav improves evidence quality mainly by reducing irrelevant context, not by reducing coverage.}

\subsection{Downstream Repair Utility}

To test whether the localization gains transfer to issue resolution, we use the context produced by each Gemma 4B localizer as the only repository context available to two fixed repair models, Gemma 4B and Qwen3.5 35B-A3B. We hold the repair prompt, generation budget, patch validation, and SWE-bench Lite test harness fixed across localization methods. Each model makes one deterministic repair attempt. Applied measures whether a generated patch can be applied to the repository, while Resolved requires it to pass the benchmark evaluation.

\input{tables/program_repair_results}

SemNav provides the best non-oracle context for both repair models (Table~\ref{tab:program_repair}). It improves the resolved rate over the strongest baseline by 4.33 points with Gemma 4B and by 8.33 points with Qwen3.5 35B-A3B, while also achieving the highest patch application rate. The benefit with both models shows that SemNav's localization gains transfer across repair backends. The larger gain with Qwen suggests that a stronger repair model can make effective use of the localized context.

SemNav remains below oracle context with both repair models, showing that better localization can further improve downstream repair. However, oracle context does not resolve every issue, so patch generation remains a separate source of error.

\takeaway{SemNav's localization gains translate into better repair performance across both backends.}

%% file: tables/program_repair_results.tex
\begin{table}[t]
    \caption{One-shot downstream repair results on SWE-bench Lite. All localization results are produced by Gemma 4B. Green parenthesized values report the fraction of the corresponding oracle resolution rate.}
    \label{tab:program_repair}
    \centering
    \begingroup
    \renewcommand{\arraystretch}{1.05}
    \scriptsize
    \setlength{\tabcolsep}{1.8pt}
    \begin{tabular}{@{}lrrrr@{}}
    \toprule
    & \multicolumn{2}{c}{Gemma 4B Repair} & \multicolumn{2}{c}{Qwen3.5 35B Repair} \\
    \cmidrule(lr){2-3}\cmidrule(l){4-5}
    Localization & Applied & Resolved & Applied & Resolved \\
    \midrule
    Agentless      & 62.67 & 14.00 {\color{green!50!black}(35.9\%)} & 75.33 & 30.67 {\color{green!50!black}(45.8\%)} \\
    CoSIL          & 73.00 & 21.67 {\color{green!50!black}(55.6\%)} & 85.00 & 42.00 {\color{green!50!black}(62.7\%)} \\
    LocAgent       & 75.33 & 23.33 {\color{green!50!black}(59.8\%)} & 86.67 & 44.00 {\color{green!50!black}(65.7\%)} \\
    mini-swe-agent & 72.00 & 19.67 {\color{green!50!black}(50.4\%)} & 83.67 & 40.00 {\color{green!50!black}(59.7\%)} \\
    RepoSearcher   & 65.00 & 15.33 {\color{green!50!black}(39.3\%)} & 77.00 & 33.00 {\color{green!50!black}(49.3\%)} \\
    \rowcolor{gray!15}[2pt][2pt]\textbf{SemNav} & \textbf{83.00} & \textbf{27.67 {\color{green!50!black}(70.9\%)}} & \textbf{91.33} & \textbf{52.33 {\color{green!50!black}(78.1\%)}} \\
    Oracle source files & 90.00 & 39.00 {\color{green!50!black}(100\%)} & 95.00 & 67.00 {\color{green!50!black}(100\%)} \\
    \bottomrule
    \end{tabular}
    \endgroup
    \end{table}

%% file: sections/related_works.tex
\section{Related Work}

\subsection{Repository-Level Issue Localization}

Repository-level issue localization follows either predefined workflows or agent-directed search. Workflow methods retrieve, filter, and rerank locations: Agentless localizes hierarchically from repository structure to fine-grained entities, SWE-Fixer combines lexical retrieval with learned localization, SweRank applies task-specific retrieval and listwise reranking, and CoSIL searches module- and function-level call graphs in stages~\cite{xia2025demystifying,xie2025swefixer,reddy2026swerank,jiang2025issuelocalization}. Agent-driven methods let an LLM acquire evidence through AST search, repository-graph traversal, action scheduling, language-server navigation, or shell tools~\cite{zhang2024autocoderover,chen2025locagent,yu2025orcaloca,zhang2026onetool,sutawika2026codescout}. Some systems retain candidate locations or repository knowledge outside the reasoning trajectory~\cite{jiang2025issuelocalization,yu2025orcaloca,ma2025alibaba}. SemNav combines broad deterministic initialization with evidence-backed agent revision.

\subsection{Repository Representation and Navigation}

Repository representations range from cross-file code fragments~\cite{zhang2023repocoder,ding2024cocomic,liang2024repofuse} to graphs of files, classes, functions, and syntax-derived relations. RepoGraph retrieves local definition and reference neighborhoods, CodexGraph supports structured graph queries, LocAgent traverses a heterogeneous entity graph, and CoSIL uses stage-specific call graphs~\cite{ouyang2025repograph,liu2025codexgraph,chen2025locagent,jiang2025issuelocalization}. RepoNavigator further provides language-server-backed definition navigation~\cite{zhang2026onetool}, while complementary methods attach natural-language descriptions or conceptual concerns to entities for semantic retrieval~\cite{athale2025knowledge,wang2025extractingconceptual}. Yet syntax-derived relations may leave targets ambiguous, and structural, semantic, and source evidence often remain separate. SemNav unifies on-demand language-server-resolved relations and issue-conditioned Semantic Cards around shared graph entities, enabling reuse of both evidence types throughout localization.